\documentclass{article}
\usepackage{times}
\usepackage{epsfig}
\usepackage{graphicx}
\usepackage{amsmath}
\usepackage{amssymb}
\usepackage{multirow}
\usepackage{algorithm} 
\usepackage{algorithmic}  
\usepackage[algo2e]{algorithm2e} 
\usepackage{dsfont}
\usepackage{bbm}
\usepackage{array}
\usepackage{comment}
\usepackage{booktabs}
\usepackage{subcaption}

\usepackage{algorithm}
\usepackage{algorithmic}

\makeatother
\usepackage[accepted]{icml2017} 

\icmltitlerunning{Reformulating Level Sets as Deep Recurrent Neural Network Approach to Semantic Segmentation}

\begin{document} 
	
	\twocolumn[
	\icmltitle{Reformulating Level Sets as Deep Recurrent Neural Network Approach \\ to Semantic Segmentation}
		
	\begin{icmlauthorlist}
	\icmlauthor{Ngan Le}{cmu}
	\icmlauthor{Kha Gia Quach}{cmu,con}
	\icmlauthor{Khoa Luu}{cmu}
	\icmlauthor{Marios Savvides}{cmu}
    \icmlauthor{Chenchen Zhu}{cmu}
	\end{icmlauthorlist}
	
	\icmlaffiliation{cmu}{Carnegie Mellon University, Pittsburgh, PA, US}
	\icmlaffiliation{con}{Concordia Univesrity, Montreal, Canada}
	
	\icmlcorrespondingauthor{Ngan Le}{thihoanl@andrew.cmu.edu}
 
	\vskip 0.3in
	]

	\printAffiliationsAndNotice{} 
	
	\begin{abstract} 
		
		Variational Level Set (LS) has been a widely used method in medical  segmentation. However, it is limited when dealing with multi-instance objects in the real world. In addition, its segmentation results are quite sensitive to initial settings and highly depend on the number of iterations. To address these issues and boost the classic variational LS methods to a new level of the learnable deep learning approaches, we propose a novel definition of contour evolution named \textit{Recurrent Level Set (RLS)}{$^1$}\footnote{$^*$Source codes will be publicly available.} to employ Gated Recurrent Unit under the energy minimization of a variational LS functional. The curve deformation process in RLS is formed as a hidden state evolution procedure and updated by minimizing an energy functional composed of fitting forces and contour length. By sharing the convolutional features in a \textit{fully end-to-end trainable framework}, we extend RLS to \textit{Contextual RLS (CRLS)} to address semantic segmentation in the wild. The experimental results have shown that our proposed RLS improves both computational time and segmentation accuracy against the classic variational LS-based method whereas the fully end-to-end system CRLS achieves competitive performance compared to the state-of-the-art semantic segmentation approaches.
		
	\end{abstract} 
	
	\begin{figure}[!t]
		\centering 
        \begin{subfigure}{1.\columnwidth}
        \centering
        \includegraphics[width=1.\columnwidth]{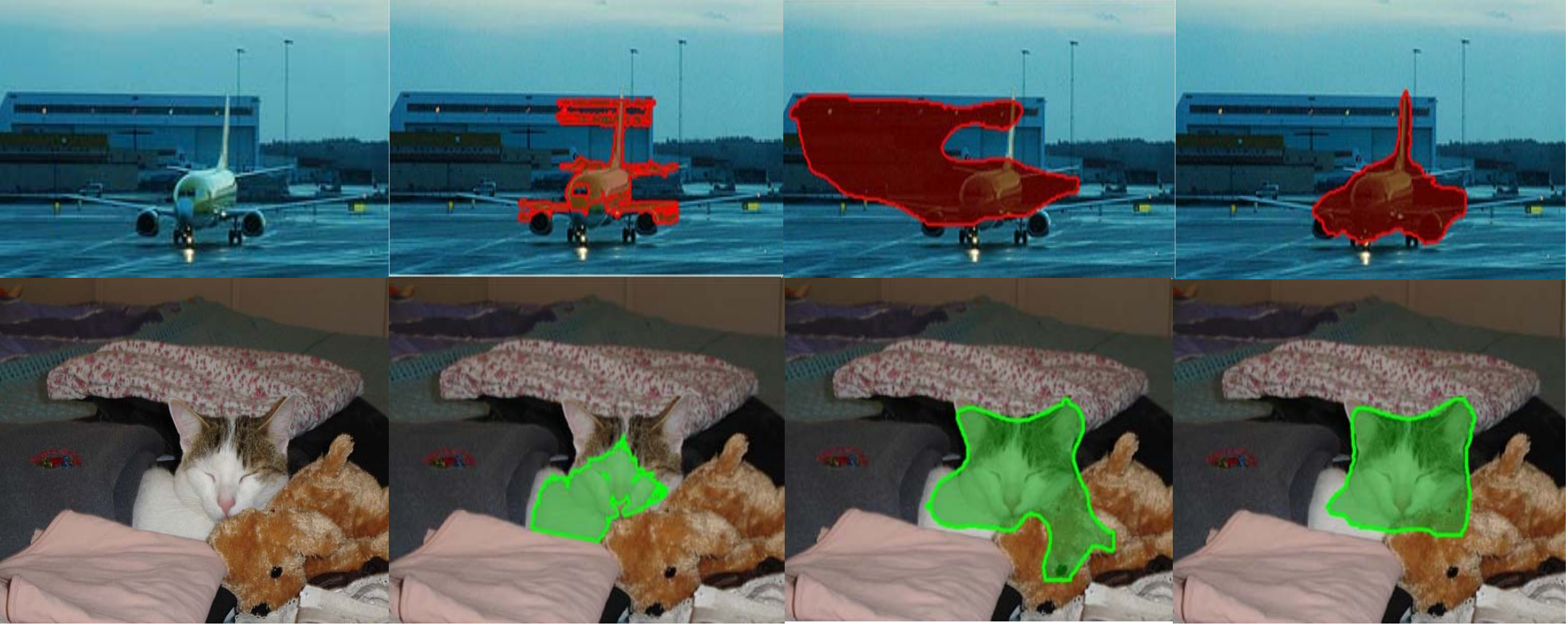}	
        \caption{Examples of \textit{object segmentation} results.}
        \label{fig:example1}
        \end{subfigure} %
		\begin{subfigure}{1.\columnwidth}
		\centering 
        \includegraphics[width=1.\columnwidth]{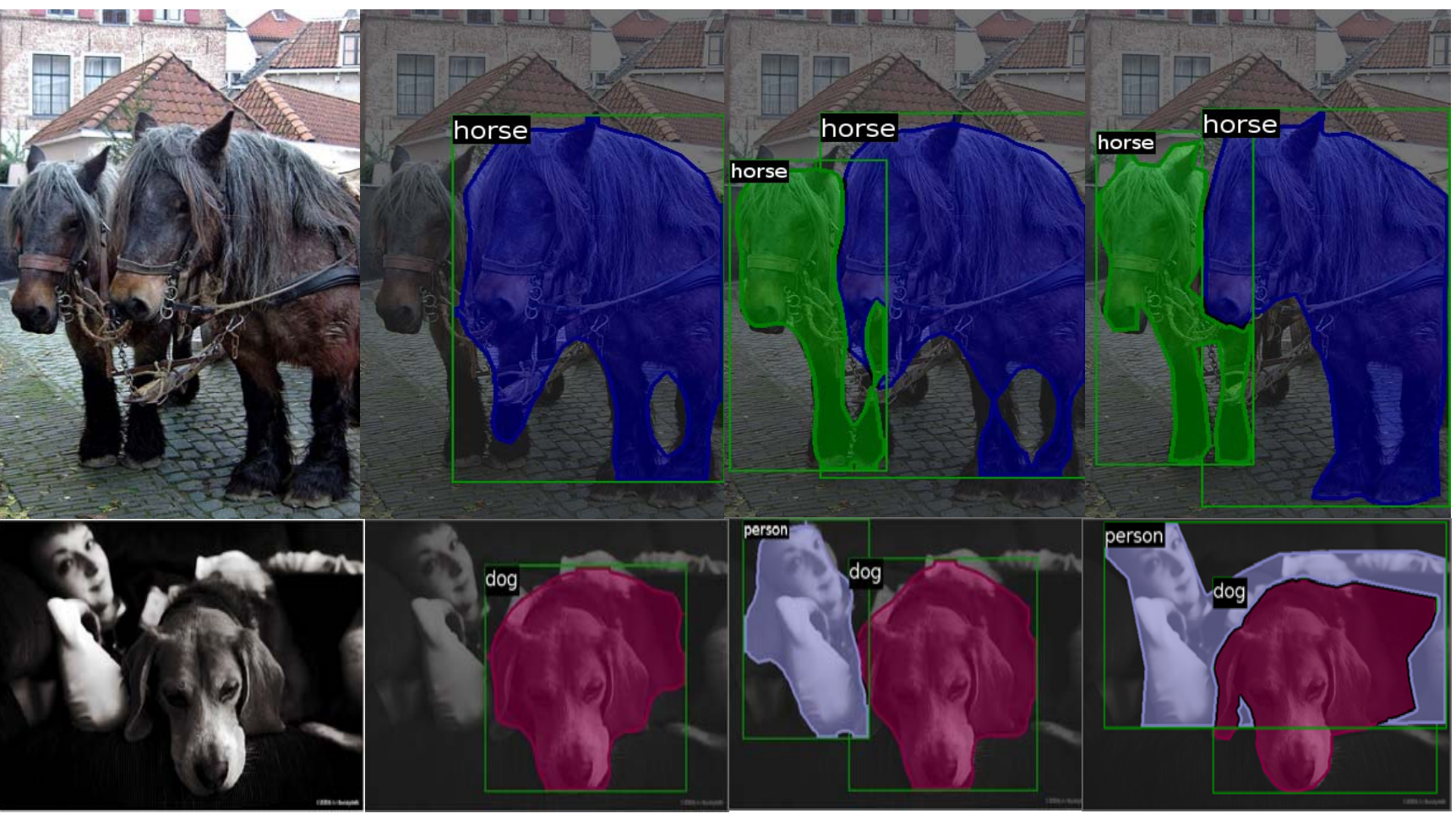}
        \caption{Examples of \textit{semantic segmentation} results.}
        \label{fig:example2}
        \end{subfigure} %
        \caption{  
        From left to right: (a) the input images, CLS \cite{chan2001active}, MNC \cite{Dai_MNC_2016} and our \textbf{RLS}; (b) the input images, MNC, our \textbf{CRLS} and ground truth images.}			\label{fig:example}
	\end{figure}
    
	\section{Introduction}
	
	Image segmentation has played an important role in various areas of computer vision and image processing. The segmentation results have been recently improved rapidly thank to deeply learned features and large-scale annotations. Among numerous segmentation methods developed in last few decades, Active Contour (AC), or Deformable Models, based on variational models and partial differential equations (PDEs), can be considered as one of the most widely used approaches in medical image segmentation. Among many AC-based approaches in the last few decades for image segmentation, variational LS methods \cite{Samson2000, Brox2006, Bae2009, Lucas2012, Huang2014, Jianbing2014} have obtained  promising performance under some constraints, e.g. resolution, illumination, shape, noise, occlusions, etc. However, the segmentation accuracy in the LS methods dramatically drops down when dealing with images collected in the wild conditions, e.g. the PASCAL Visual Object Classes (VOC) Challenge \cite{Everingham15}, the Microsoft Common Objects in COntext (MS COCO) \cite{MSCOCO} database, etc. Meanwhile, the recent advanced deep learning based segmentation approaches \cite{Hariharan2014_SDS, Dai_MNC_2016, Girshick2015, Long_2015_fullNN, Zheng_2015_CRF} have achieved the state-of-the-art performance on dealing with multi-instance object segmentation on these databases. Indeed, the limitations of the LS approaches can be observed as follows. Firstly, the LS methods have a large handicap in capturing variations of real-world objects due to the nature of only using pixel values in these methods. Secondly, the LS methods are unable to memorize and to fully infer target objects since they do not have any learning capability. Thirdly, the LS based methods are very limited in segmenting multiple objects with semantic information. Furthermore, the segmentation performance by the LS methods is quite sensitive to numerous pre-defined parameters including initial contour, number of iterations.
	
	To address the aforementioned problems, this paper presents a novel look of LS methods under the view point of a deep learning approaches, i.e. Recurrent Neural Network (RNN) and Convolutional Neural Networks (ConvNet). The classic LS (CLS) segmentation formulation is now redefined as a recurrent learning framework, named  \textbf{Recurrent Level Set (RLS)}, thank to its evolution properties, i.e. stability and the growth of oscillations in the models. Compare to CLS method \cite{chan2001active}, RLS is more robust and productive when dealing with images in the wild thanks to learning ability as shown in Fig.~\ref{fig:example1}. The proposed RLS is designed as a trainable system thus it is easily extended to a fully end-to-end system, named \textbf{Contextual RLS (CRLS)} to efficiently incorporate with other deep learning frameworks for handling the semantic segmentation in the wild. Compared to other ConvNet based segmentation methods \cite{Hariharan2014_SDS,  Girshick2015, Long_2015_fullNN, Dai_MNC_2016}, our proposed fully end-to-end trainable deep networks CRLS inherits all the merits of the LS model to enrich the object curvatures and the ConvNet model to encode powerful visual representation. Indeed, the proposed CRLS is able to learn both object contours via the LS energy minimization and visual representation via the sharing deep features as shown in some examples in Fig.~\ref{fig:example2}. 
	
	To the best of our knowledge, this is the first study on formulating LS-based method under a learnable framework. The contribution of our work can summarized as follows: (1) bridge the gaps between the pure image processing variational LS methods and learnable computer vision approaches; (2) propose a new formulation of curve evolution under an end-to-end deep learning framework by reforming the optimization process of CLS as a current network, named RLS; (3) The proposed RLS framework is formed as a building block which is easily incorporated with other existing deep modules to provide a robust deep semantic segmentation, named CRLS. 	
	\section{Related Work}
	
	\subsection{Level Sets based Approaches}
	
	The key ideas behind the AC for image segmentation is to start with an initial guess boundary represented in a form of closed curves i.e. contours $C$. The curve is then iteratively modified by applying shrink or expansion operations and moved by image-driven forces to the boundaries of the desired objects. The entire process is called contour evolution, denoted as $\frac{\partial C}{\partial t}$.     
	There are two kinds of AC models, regarding the LS method with respect to image segmentation, i.e. edge-based and region-based. Edge-based LS uses an edge detector to extract the boundaries of sub-regions \cite{Caselles1993}. However, the edge-based LS suffers some weaknesses, e.g. sensitivity to noise and some level of prior knowledge is still required. Region-based LS, or variational LS, was later proposed to overcome these limitations by measuring the uniformity property within each subset instead of searching geometrical boundaries. Chan-Vese's (CV) model \yrcite{chan2001active} is one of the most successful approaches in this category. 
	 
    To overcome the limitation of CLS being binary-phase segmentation, Samson et al. \yrcite{Samson2000} associates a LS function with each image region, and evolves these functions in a coupled manner. Later, Brox \& Weickert \yrcite{Brox2006} performs hierarchical segmentation by iteratively splitting previously obtained regions using the CLS.  Lucas \yrcite{Lucas2012} suggested using a single LS function to perform the LS evolution for multi-region segmentation. It requires managing multiple auxiliary LS functions when evolving the contour, so that no gaps/overlaps are created. Bae \&  Tai \yrcite{Bae2009} proposed to partition an image into multiple regions by a single, piecewise constant LS function, obtained using either augmented Lagrangian optimization, or graph-cuts. Recently, Dubrovina et al. \yrcite{Dubrovina2015} has developed an multi-region segmentation with single LS function. In addition to multi region segmentation problem, optimization \cite{Huang2014, Jianbing2014, Zhang2015}, and shape prior \cite{shape_ac_PR2016} have been also considered. 
	
	\subsection{Semantic Segmentation Approaches}
	
	Semantic segmentation refers to associating one of the classes to each pixel in an image. This problem has played an important role in many research areas and has attracted numerous studies recently. In recent years, deep learning techniques have become ubiquitous in semantic segmentation. One of the first studies to apply ConvNet to semantic segmentation was \cite{farabet2013learning}, which stacked encompassing windows from different scales to serve as context. Instead, Long et al. \yrcite{Long_2015_fullNN} proposed to use fully connected convolutional networks to utilize model complexity. Shuai et al. \yrcite{shuai2015integrating} combined parametric and non-parametric techniques to model global order dependencies to provide more information and context. Lately, recurrent models have started to gain popularity. For example, Pinheiro et al. \yrcite{pinheiro2014recurrent} proposed to pass an input image through a ConvNet multiple times in sequence \emph{i.e.}. The output of the ConvNet is fed into the same ConvNet again. Zheng et al. \yrcite{Zheng_2015_CRF} modeled a CRF as a neural network that is iteratively applied to an input, thereby qualifying as a recurrent model. Inference is done through convergence of the neural network output to a fixed point. Though these methods are a form of recurrent models, the capacity to capture weaker long range dependencies is limited due to lack of explicit sequence modeling and sole dependence on filter based modeling. Recently, Pinheiro et al. \yrcite{DeepMask} has presented a new approach to robustly segment objects from given images. In this system, the discriminative ConvNet is used to generate object proposals. Given an image patch, the training objective function includes two tasks, i.e. class-agnostic segmentation and likelihood of the patch being centered on a full object. This work and their refined version \cite{SharpMask} have achieved the top segmentation performance in the field.

	\section{Recurrent Level Sets (RLS): Model and Learning}
	\label{sec:RLS}
	
	In this section, we first review the classic Level Set (CLS) model proposed by Chan-Vese \yrcite{chan2001active} in Subsec.~\ref{subsec:CV}. We then details the proposed Recurrent Level Sets (RLS) which inherits all the merits of LS and Gate Recurrent Unit (GRU) \cite{Cho2014} to formulate a new definition of curve evolution in Subsec.~\ref{subsec:RLS}. Finally, the learning process of our RLS model is described in Subsec.~\ref{subsec:RLS_learning}.
	
\subsection{Classic Level Set (CLS) - Revisited}
\label{subsec:CV}	
	One of the most successful CLS models was proposed by Chan \& Vese \yrcite{chan2001active}. In their model, the energy minimization problem is defined as in \eqref{eq:Chan_Vese_3}.
\begin{equation}
\begin{alignedat}{2}
& \min_{c_1, c_2, \varphi}  \mu \int_{\Omega}{H(\varphi)dxdy} + \nu \int_{\Omega}{\delta(\varphi)|\nabla\varphi| dxdy} & (1)\\
& + \int_{\Omega} \left( \lambda_1 {|\textbf{I}-c_1|^{2}H(\varphi)  + \lambda_2 |\textbf{I}-c_2|^{2}(1-H(\varphi))} \right) & dxdy \nonumber
\end{alignedat}
\label{eq:Chan_Vese_3}    
\end{equation}
\setcounter{equation}{1}
where $\textbf{I}$ denotes an input image, $C$ is the contour and $\varphi$ is the zero LS defined as ${C = \{(x,y) : \varphi(x,y) = 0\}, \forall (x,y) \in \Omega}$, where $\Omega$ denotes the entire domain of an image $\textbf{I}_{x,y}$. The parameters $\mu, \nu, \lambda_1, \lambda_2$ are positive parameters.  
The length and the area inside the contour, Heaviside function $H_{\epsilon} ({\cdot})$ and its derivative $\delta _{\epsilon} ({\cdot}) = H'_{\epsilon}(\cdot)$, are defined and regularized as 
$H_{\epsilon} ({t}) = \dfrac{1}{2} \left( 1+\frac{2}{\pi}\tan^{-1} \left( \frac{{t}}{\epsilon}\right) \right)$, and 
$\delta _{\epsilon} ({t})=\frac{\epsilon}{\pi (\epsilon ^2+{t}^2)}$, respectively.
The average values of the inside and outside of the contour, $c_1$ and $c_2$, are defined as follows:
\begin{equation}
\small
\begin{split}
c_1  = \frac{\int_{\Omega}{\textbf{I}_{x,y} H(\varphi_t) dxdy}}{\int_{\Omega} H(\varphi_t)dxdy} \text{ , }
c_2 =  \frac{\int_{\Omega}{\textbf{I}_{x,y}(1-H(\varphi_t))dxdy}}{\int_{\Omega}(1-H(\varphi_t))dxdy}
\end{split}
\label{eq:c}
\end{equation}

Given fixed $c_1$ and $c_2$, the gradient descent respecting to $\varphi$
\begin{eqnarray}
\small
\begin{split}
\frac{\partial \varphi_t}{\partial t} & = \delta_\epsilon(\varphi_t[\nu\kappa(\varphi_t - \mu 
& - \lambda_1(\textbf{I} - c_1)^2 + \lambda_2(\textbf{I} - c_2)^2 ])
\label{eq:evolution}
\end{split}
\end{eqnarray}
where 
the curvature $\kappa(\varphi_t) = -div\left ( \frac{\nabla \varphi_t}{\left | \nabla \varphi_t \right |}\right )$ is given by
\begin{equation}
\small
\frac{\partial_{xx}\varphi_{t}\partial_{y}^2\varphi_t - 2\partial_{x}\varphi_t\partial_{y}\varphi_t\partial_{xy}\varphi_{t} + \partial_{yy}\varphi_{t}\partial_{x}^2\varphi_t}{\left ( \partial_{x}^2\varphi_t + \partial_{y}^2\varphi_t\right )^{3/2}}
\label{eq:kappa}
\end{equation}
where $\partial_{x}\varphi_{t}, \partial_{y}\varphi_{t}$ and $\partial_{xx}\varphi_{t}, \partial_{yy}\varphi_{t}$ are the first and second derivatives of $\varphi_t$ with respect to $x$ and $y$ directions.

From this point, we redefine the curve updating in a time series form for the LS function $\varphi_t$ as in Eqn.~\eqref{eq:varphi_update}.
\begin{equation}
\varphi_{t+1} = \varphi_{t} + \eta\frac{\partial \varphi_t}{\partial t}
\label{eq:varphi_update}
\end{equation}
The LS at time $t+1$ depends on the previous LS at time $t$ and the curve evolution $\frac{\partial \varphi_t}{\partial t}$ with a learning rate $\eta$.

\begin{figure}[!t]
\centering \includegraphics[width=1.0\columnwidth]{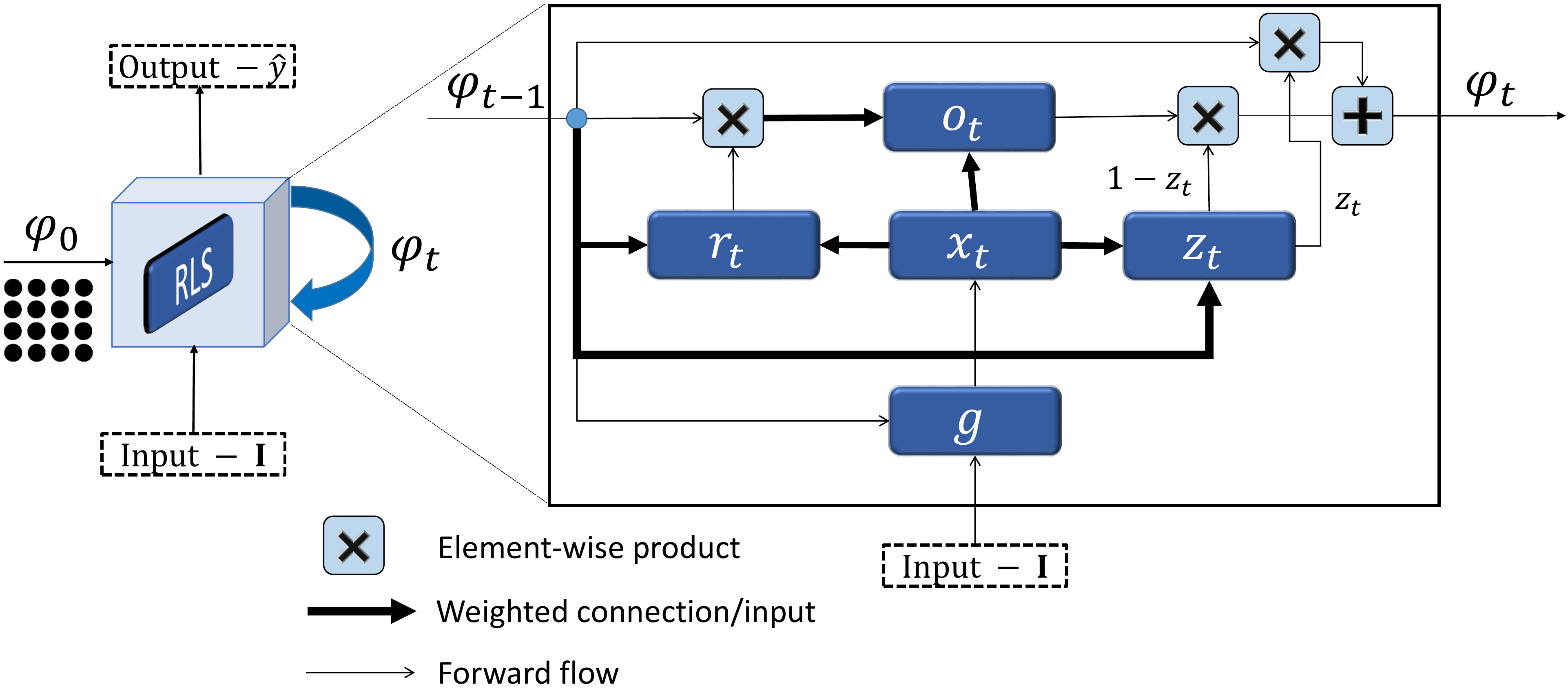}
		\caption{The proposed RLS network for curve updating process under the sequential evolution and its forward computation of curve evolution from time $t-1$ to time $t$.}
\label{fig:RLS}
\end{figure}

\begin{figure*}[!t]
	\centering \includegraphics[width=2.0\columnwidth]{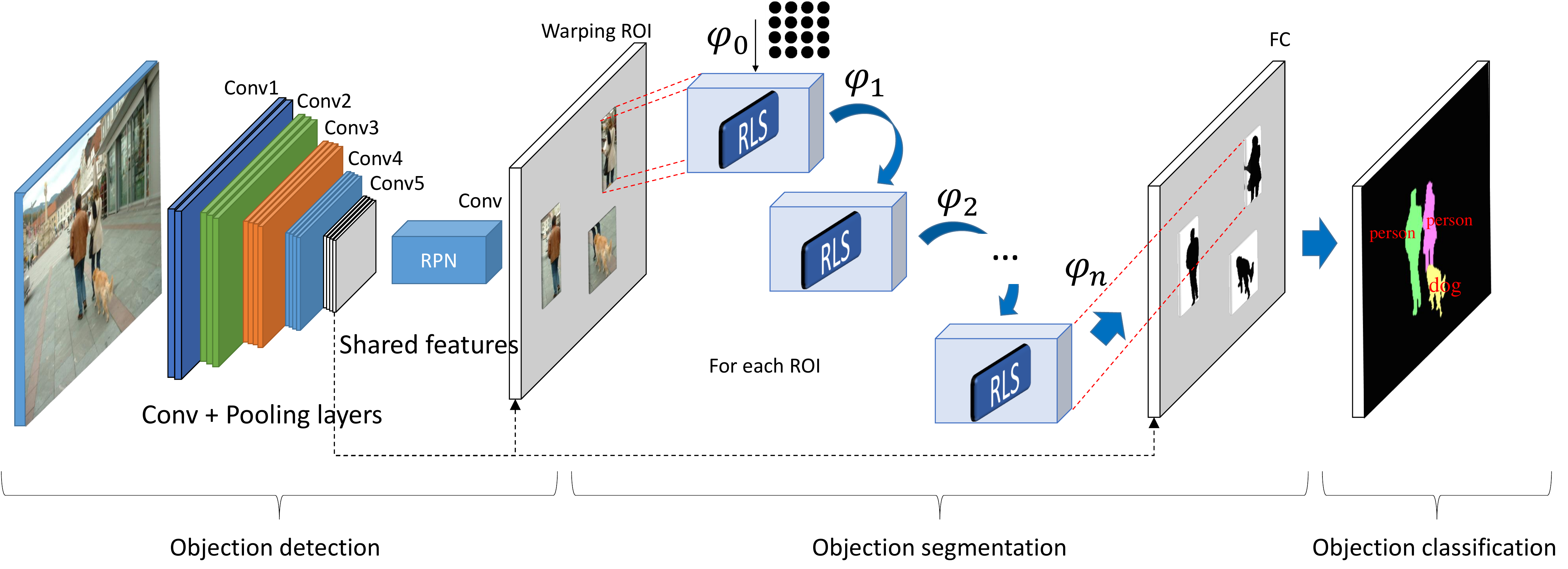}
		\caption{The flowchart of our proposed CRLS for semantic segmentation with three tasks: object detection by Faster R-CNN \cite{Ren_2015_FRCNN}, object segmentation by RLS and classification}
	\label{fig:flowchart}
\end{figure*}
	
\subsection{Recurrent Level Set (RLS) - Proposed}
\label{subsec:RLS}
In this section, we take the CLS evolution introduced in \cite{chan2001active} as an instance to demonstrate the idea of how to reformulate LS as an end-to-end trainable recurrent framework, named RLS. However, the proposed RLS can be applied to reform any LS approach once it successfully reforms CLS model because they share similar properties of curves moving over time. In our proposed RLS approach, the recurrent units work in the same fashion as GRU. In order to reformulate CLS in a deep learning framework, we first study how CLS communicates with a recurrent network. 
	
	In the CLS framework, the curve updating is presented in a time series form as in Eqn.~\eqref{eq:varphi_update}. The zero LS function at time $t+1$ depends on the previous zero LS at time $t$ and the curve evolution $\frac{\partial \varphi}{\partial t}$ with a learning rate $\eta$. Compared to recurrent network, the curve updating in Eqn.\eqref{eq:varphi_update} can be presented as a hidden process under time series. This advanced relationship gives us a big favor when reforming the CLS under the recurrent framework.
	
	However, the most difficult part of reformulating CLS as recurrent network is data configuration. Recurrent network works on sequence data while both the input and the output of the CLS approach are single images. The critical question is \textit{how to generate sequence data from a single image}. Notably, there are two inputs used in CLS. One is an input image $\textbf{I}$ and the other is an initial LS function $\varphi_0$, updated using Eqn.~\eqref{eq:varphi_update}. The LS function $\varphi_t$ at time $t$ follows the curve evolution process defined by Eqn.~\eqref{eq:evolution}. During the curve evolution, the input image $\textbf{I}$ is used to compute the energy force that helps to drive the curve. Based on these observations, the sequence data $\textbf{x}_t$ in our proposed RLS is generated from the single input image $\textbf{I}$ through a function $g$ and under the control of zero LS function $\varphi$ as shown in Eqn.~\eqref{eq:x_i}. The proposed RLS in folded mode is given in Fig.~\ref{fig:RLS} (left) where the input of the network is defined as the same as the CLS model, namely, an input image $\textbf{I}$ and an initial zero LS function $\varphi_o$. The zero LS function is initialized in a similar fashion as in \cite{chan2001active} via checkerboard function (See Fig. \ref{fig:RLS}-Left). In our proposed RLS, the curve evolution from $\varphi_{t-1}$ at time $t-1$ to the next step $\varphi_{t}$ at time $t$ is illustrated in Fig.\ref{fig:RLS}(right) where  $\varphi_{t}$ depends on both  $\varphi_{t-1}$ and the input image $\textbf{I}$.
	\begin{equation}
    \small
		\begin{split}
			\textbf{x}_{t}  & = g(\textbf{I}, \varphi_{t-1}) = \kappa(\varphi_{t-1}) - \textbf{U}_g(\textbf{I} - c_1)^2 + \textbf{W}_g(\textbf{I} - c_2)^2
			\label{eq:x_i}
		\end{split}
	\end{equation}
	In Eqn.~\eqref{eq:x_i},  $c_1$ and $c_2$ are average values of inside and outside of the contour presented by the zero LS function $\varphi_{t}$ and defined in Eqn.~\eqref{eq:c}. $\kappa$ denotes the curvature and defined in Eqn.~\eqref{eq:kappa}. $\textbf{U}_g$ and $\textbf{W}_g$ are two matrices that control the force inside and outside of the contour.
	
	In our proposed RLS, the zero LS function $\varphi_t$ is treated as a vector of activation/hidden units in a recurrent network, i.e. vectorized zero LS function $\varphi_t$. It is updated using the updated gate $\textbf{z}_t$, the candidate memory content $\textbf{o}_t$ and the previous activation unit $\varphi_{t-1}$ as the rule given in Eqn.~\eqref{eq:rls_phi}.
	\begin{equation}
    \mathbf{\varphi}_{t} =  \textbf{z}_t \odot \mathbf{\varphi}_{t-1}  + (1-\textbf{z}_t) \odot \textbf{o}_{t}\\    
		\label{eq:rls_phi}
	\end{equation}
	where $\odot$ denotes an element-wise multiplication. The update gate $\textbf{z}_t$, which controls how much of the previous memory content is to be forgotten and how much of the new memory content is to be added is defined as in Eqn.~\eqref{eq:z}. 
	\begin{equation}
		\textbf{z}_t  =  \sigma(\textbf{U}_z\textbf{x}_t + \textbf{W}_z\varphi_{t-1} + \textbf{b}_z)\\
		\label{eq:z}
	\end{equation}
	where $\sigma$ is a sigmoid function and $\textbf{b}_z$ is the update bias. The RLS, however, does not have any mechanism to control the degree to which its state is exposed, but exposes the whole state each time. The new candidate memory content $\textbf{o}_t$ is computed 
as in Eqn.~\eqref{eq:h_tilde}.
	\begin{equation}
		\textbf{o}_t = tanh(\textbf{U}_{o}\textbf{x}_t + \textbf{W}_{o}(\varphi_{t-1} \odot \textbf{r}_t) + \textbf{b}_{o})
		\label{eq:h_tilde}
	\end{equation}
	where $\textbf{b}_{o}$ is the hidden bias. 
The reset gate $\textbf{r}_t$ is computed similarly to the update gate as in Eqn.~\eqref{eq:r}.
	\begin{equation}
		\textbf{r}_t = \sigma(\textbf{U}_r\textbf{x}_t + \textbf{W}_r\varphi_{t-1} + \textbf{b}_r)
		\label{eq:r}
	\end{equation}
	where $\textbf{b}_r$ is the reset bias. When $\textbf{r}_t$ is close to 0 (off), the reset gate effectively makes the unit act as if it is reading the first symbol of an input sequence, allowing it to forget the previously computed state. 
The output $\textbf{O}$ is computed from the current hidden states $\varphi_t$ 
and then a softmax function is applied to obtain foreground/background segmentation $\hat{\textbf{y}}$ given the input image 
as follows, 
\begin{equation} \label{eq:y_hat}
 \hat{\textbf{y}} = \mathrm{softmax}(\textbf{O}) = \mathrm{softmax}(\textbf{V} \varphi_t + \mathbf{b})
\end{equation}
where $\textbf{V}$ is weighted matrix between hidden state and output. The proposed RLS model is trained in an end-to-end framework and its  learning processes of the forward pass and the backward propagation are described as follows.

\subsection{RLS Learning}
\label{subsec:RLS_learning}
The Back Propagation Through Time method is used to train the parameter set $\theta = \{$ $\textbf{U}_g$, $\textbf{W}_g$, $\textbf{U}_z$, $\textbf{W}_z$, $\textbf{U}_r$, $\textbf{W}_r$, $\textbf{U}_o$, $\textbf{W}_o$,$\textbf{V}$ $\}$ and propagate error backward through time${^2}$\footnote{$^2$The derivatives of all parameters are detailed in the supplementary material.}. 
We apply RMS-prop \cite{dauphin1502rmsprop} with momentum $\rho_m = 0.9$.
This optimizer minimize the following cross entropy loss function,
\begin{equation}
L_1(\hat{\textbf{y}}, \textbf{y}, \theta) = - \sum_{n=1}^{N} { \sum_{k = 1}^{K} \textbf{y}_{nk}log( \hat{\textbf{y}}_{nk})}
\end{equation}
where $N$ is the number of pixels, $K$ is the number of classes ($K=2$), since we only segment foreground/background. $\mathbf{\hat{y}}_{nk}$ is our predictions, and $\textbf{y}_{nk}$ is the ground truth.  
Normalized and smoothed gradients of similar size for all weights are used such that even weights with small gradients get updated. This also helps to deal with vanishing gradients.
We train our model using a decaying learning rate starting from $\eta = 10^{-3}$ and dividing by half every 200 epochs asymptotically toward $\eta = 10^{−5}$ for 5000 epochs in total.

We summarize the proposed building block RLS in Algorithm~\ref{alg:RLS_alg}.

\begin{algorithm}
	\caption{The proposed building block RLS}
	\begin{algorithmic}
		\STATE \textbf{Input:} Given an image $\textbf{I}$, an initial level set function $\varphi_0$, time step $T$, learning rate $\eta$, initial parameters $\theta = $ $(\textbf{U}_z$, $\textbf{W}_z$, $\textbf{b}_z$, $\textbf{U}_r$, $\textbf{W}_r$, $\textbf{b}_r$  $\textbf{U}_{o}$, $\textbf{W}_{o}$, $\textbf{b}_{o}$, $\textbf{V}$, $\textbf{b}_V)$      
		\FOR {each epoch}		
		\STATE Set $\varphi = \varphi_0$
        \FOR {t = 1 : T}
		\STATE{Generate RLS input $\textbf{x}_t$:}
		{$\textbf{x}_{t}  \leftarrow g(\textbf{I}, \varphi_{t-1})$}
		\STATE{Compute update gate $\textbf{z}_t$, reset gate $\textbf{r}_t$ and intermediate hidden unit $\tilde{\textbf{h}}_t$:}
		\STATE {$\textbf{z}_t  \leftarrow  \sigma(\textbf{U}_z\textbf{x}_t + \textbf{W}_z\varphi_{t-1} + \textbf{b}_z)$}
		\STATE {$\textbf{r}_t  \leftarrow \sigma(\textbf{U}_r\textbf{x}_t + \textbf{W}_r\varphi_{t-1} + \textbf{b}_r)$}
		\STATE {$\textbf{o}_t \leftarrow tanh(\textbf{U}_{o}\textbf{x}_t + \textbf{W}_{o}\left(\varphi_{t-1} \circ \textbf{r}_t\right) + \textbf{b}_{o})$}
		\STATE{Update the zero LS $\varphi_t$: $\varphi_t \leftarrow (1-\textbf{z}_t)\textbf{o}_t + \textbf{z}_t\varphi_{t-1}$}
        \ENDFOR
		\STATE {Compute the loss function $L$: $L \leftarrow -\sum_n{\textbf{y}_n log{\hat{\textbf{y}_n}}}$}
		\STATE {Compute the derivate w.r.t. $\theta$: $\nabla \theta \leftarrow \frac{\partial L}{\partial \theta}$}
		\STATE {Update $\theta$: $\theta \leftarrow \theta + \eta \nabla\theta$}		
		\ENDFOR
	\end{algorithmic}
    \label{alg:RLS_alg}
\end{algorithm}

\section{Contextual Recurrent Level Sets (CRLS): Model, Inference, and Learning}
\label{sec:CRLS}

In this section, we introduce our Contextual Recurrent Level Sets (CRLS) for semantic object segmentation which is an extension of our proposed RLS model to address the multi-instance object segmentation in the wild. The output of our CRLS is multiple values (each value is corresponding to one object class) instead of two values (foreground and background) as in RLS. The entire proposed CRLS modle is first introduced in Sec.\ref{subsec:CRLS_model}. Inference and training process are then described in Sec.\ref{subsec:CRLS_inference} and \ref{subsubsec:CRLS_training},

\subsection{Model constructing}
\label{subsec:CRLS_model}

Our proposed CRLS inherits the merits of RLS and Faster-RCNN \cite{Ren_2015_FRCNN} for semantic segmentation which  simultaneously performs three tasks, i.e. detection, segmentation and classification in a fully end-to-end trainable framework as shown in Fig.~\ref{fig:flowchart}. In the proposed CRLS, the current task depends on the output of the previous one, e.g. the segmentation task that separates the foreground out of the background takes the shared deep features and the bounding boxes from the earlier detection task as its inputs. Similarly, the classification task then takes the segmenting results together the shared deep features as its inputs.

\textbf{Object detection via Region-based Convolutional Neural Networks:} 
One of the most important approaches to the object detection and classification problems is the generations of Region-based Convolutional Neural Networks (R-CNN) methods \cite{Girshick2015, girshick2015fast, Ren_2015_FRCNN}. 
Aiming to design a \textit{fully end-to-end trainable} framework, we adapt the Region Proposal Network (RPN) introduced in Faster R-CNN \cite{Ren_2015_FRCNN} to predict the object bounding boxes and the objectness scores. By sharing the convolutional features of a deep VGG-16 network \cite{simonyan2014very}, the whole system is able to perform both detection and segmentation computation efficiently.
The shared features are obtained by 13 convolution layers where each convolution layer is followed by a ReLU layer but only four pooling layers are placed right after the convolution layer to reduce the spatial dimension. As shown in Fig.~\ref{fig:flowchart}, the first section of CRLS is divided into 5 main components, i.e. \textit{conv1}, \textit{conv2}, \textit{conv3}, \textit{conv4} and \textit{conv5}, by those pooling layers. Each component consists of 2 -- 3 convolution layers. Then, the RPN consists of a $3\times3$ convolutional layer reducing feature dimensions and two consecutive $1\times1$ convolutional layers predicting \textit{object's locations} and \textit{object's presenting scores}. The location regression is with reference to a series of pre-defined boxes, or anchors, at each location.

\textbf{Object segmentation via CRLS}:
For each predicted box, we first extract a fixed-size ($m \times m$) deep feature (conv5) via \textit{ROI warping layer} \cite{Dai_MNC_2016} which crops and warps a region on the feature map into the target size by interpolation. The extracted features are passed through the proposed RLS together with a randomly initial ${\varphi_0}$ to generate a sequence input data $\textbf{x}_t$ based on Eqn.~\eqref{eq:x_i}. The curve evolution procedure is performed via LS updating process given in Eqns. \eqref{eq:evolution} and \eqref{eq:varphi_update}. This task outputs a binary mask as given in Eqn.~\eqref{eq:y_hat} sized $m \times m$ and parameterized by an $m^2$ dimensional vector, where $ m = 21$.

\textbf{Object classification via fully-connected network}:
For each box candidate, a feature representation is firstly extracted by RoI-pooling from the shared convolutional features inside the box region. Then it is masked by the segmenting mask prediction, which pays more attention on the foreground features. After that the masked feature goes through two fully-connected layers, resulting a mask-based feature vector for classification. 
The mask-based feature vector is concatenated with another box-based feature vector to build a joined feature vector. Finally, two fully-connected layers are attached to the joined feature and each gives class scores and refined bounding boxes, respectively.

\subsection{Inference}
\label{subsec:CRLS_inference}

Given an image, the top-scored 300 ROIs are first chosen by RPN proposed boxes. Non-maximum suppression (NMS) with an intersection-over-union (IoU) threshold 0.7 is used to filter out highly overlapping and redundant candidates. Then, for each ROIs, we apply our proposed RLS on top to get its foreground mask. From the ROIs and segmenting mask, the category score of each object instance is predicted via two fully-connected layers followed by a soft-max layer. 
Overlapping object instances (IoU $\geq 0.5$) are combined by averaging on a per-pixel basis, weighted by their classification scores and binarized to form the final output mask. This post-processing is performed for each category independently.

\subsection{Learning}
\label{subsubsec:CRLS_training}

During training, the share convolutional features are obtained by 13 convolution layers that initialized using the pre-trained VGG-16 model on ImageNet \cite{simonyan2014very}. 
An ROI is considered positive if its box IoU with respect to the nearest ground truth object is larger than 0.5. There are three loss scores assigned for each ROI (1) a bbox regression loss, (2) softmax segmentation to get foreground mask by our proposed RLS (3) softmax regression to classify loss over $C$ categories, where $C$ is the number of classes in the database.

The proposed CRLS semantic segmentation is implemented using Caffe environment \cite{jia2014caffe}. Training images are resized to rescale their shorter side to 600 and use SGD optimization. On PASCAL VOC \cite{pascal-voc-2012}, we perform 32k and 8k iterations at learning rates of 0.001 and 0.0001, respectively. On the MSCOCO \cite{MSCOCO}, we perform 180k and 20k iterations at learning rates of 0.001 and 0.0001, respectively.

\section{Experimental Results}

We conduct two experiments corresponding to the object segmentation by the proposed RLS method and the semantic segmentation using the proposed CRLS system. All the experiments are performed on a system of Core i7-5930k @3.5GHz CPU, 64.00 GB RAM with a single NVIDIA GTX Titan X GPU.

\subsection{RLS Object Segmentation}
\label{subsec:exp_cv}

This section compares our object segmentation RLS against Chan-Vese's LS \cite{chan2001active} and a simple Neural Network (NN) with two fully-connected layers with a ReLU layer in between (FCN). The input for object segmentation for all methods is raw pixels of an image to be segmented.
The experiments are validated on synthetic images, medical images and natural images collected in the wild. The input images are resized to $64 \times 64$, thus, the number of units in fully-connected layers and hidden cell of GRU is 4096. 
The synthetic and medical database containing 720 images are artificially created from \cite{chan2001active, Li2008} with different kinds of degradation and various affine transformations. In this dataset, we use 360 images for training and 360 images for testing. Meanwhile, the real images are collected from Weizmann \cite{AlpertGBB07} databases which has 4,700 images augmented using different kinds of noise and various affine operations (such as rotation, translation, scale, and flip). In total, we have about 6000 images for training and testing. We then split into training and testing set equally. 

\begin{figure}[!t]
	\centering \includegraphics[width=1\columnwidth]{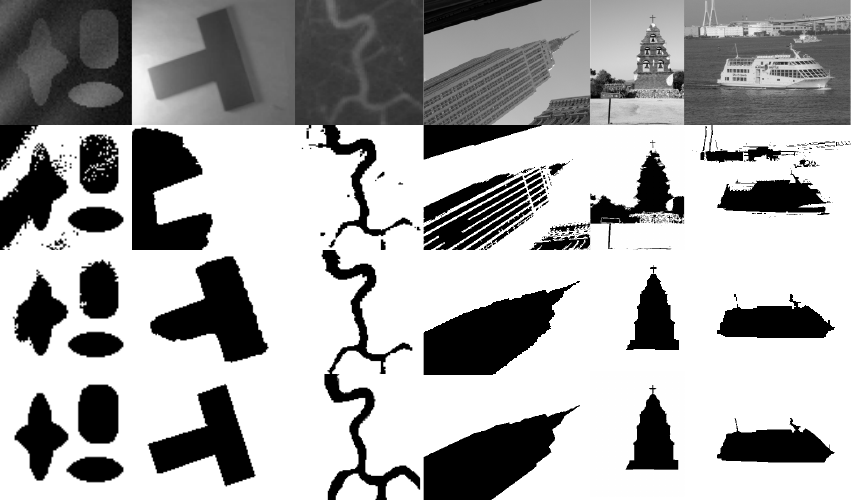}
		\caption{Object segmentation results using the images from \cite{Li2008, chan2001active, AlpertGBB07}. $1{st}$ row: input images, $2^{nd}$ row: segmentation by CLS \cite{chan2001active}, $3^{rd}$ row: segmentation our RLS, $4^{st}$ row: groundtruth.}
	\label{fig:compare_synthetic}
\end{figure}

Fig.~\ref{fig:compare_synthetic} shows some segmentation results using the baseline CLS model \cite{chan2001active} and our RLS on the synthetic, medical and natural databases. The best results from CLS's model are given in the second row. The third row shows our RLS segmentation results.  The last row shows the ground truth. In each instance, the segmentation result is given as a black/white image. The average F-measure on the real image test set obtained by CLS model, FCN and our proposed RLS are reported as in Tables \ref{tb:F_measure} with two separated groundtruth versions (GT1 and GT2) provided by two different people. 
In terms of speed, CLS method consumes 13.5 seconds on average to process one image with original size whereas RLS takes 0.008 seconds and FCN takes 0.001 seconds on average testing time to segment an object. RLS achieves the best segmentation performance in this experiment on both groundtruth annotated by two different people. From this experiment, we can see that both NN and RNN can learn and memorize the segmentation of an object in the given image. The reason why recurrent neural network perform the segmentation task better is that it actually models the curve evolution of LS much better and have the ability of fine-tuning the final segmented shape.

\begin{table}[!t]
\centering
\caption{Average F-measure (FM) and testing time obtained by CV's model, FCN and our proposed RLS with two different ground truth (GT1 and GT2) across Weizmann database}
\label{tb:F_measure}
\begin{tabular}{lccc}
\toprule[\headrulewidth]
\textbf{Methods}       & \textbf{FM (GT1)} & \textbf{FM (GT2)} &  \textbf{Testing Time}\\ \midrule 
CV    	&	88.51 &	87.51 & 13.5(s)\\ 
FCN & 93.30 & 93.26 & \textbf{0.001} (s) \\ 
RLS (Ours) &	\textbf{99.16}	&	\textbf{99.17}    & 0.008 (s)    \\ 
\bottomrule
\end{tabular}
\end{table}

\subsection{CRLS Semantic Segmentation}
\label{subsec:exp_voc}

\begin{figure*}[!t]
	\centering \includegraphics[width=2.0 \columnwidth]{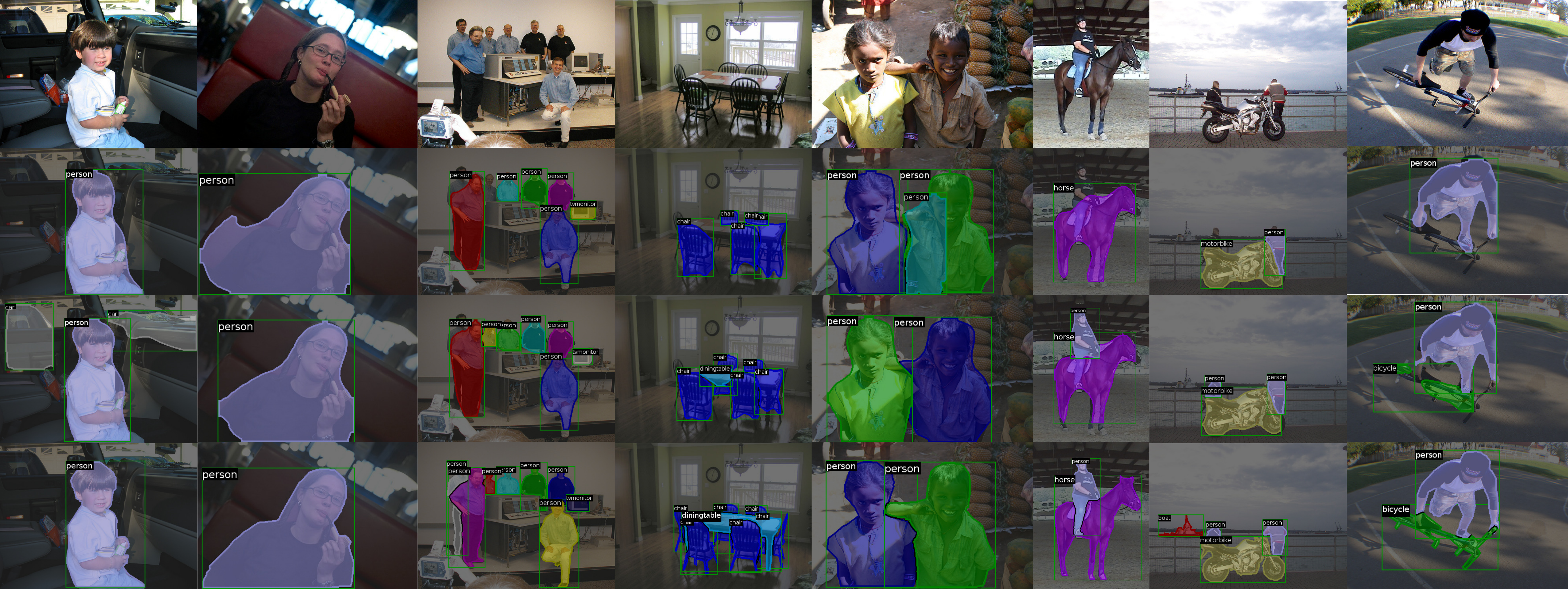}
		\caption{Some examples of semantic segmentation on PASCAL VOC 2012 database. Each column has four images: the input image ($1^{st}$ row), MNC \cite{Dai_MNC_2016}, our semantic segmentation CRLS ($2^{nd}$ row) and the ground truth ($3^{rd}$ row). (Best viewed in color)}
	\label{fig:voc2012}
\end{figure*}

\begin{figure*}[!t]
	\centering \includegraphics[width=2.0 \columnwidth]{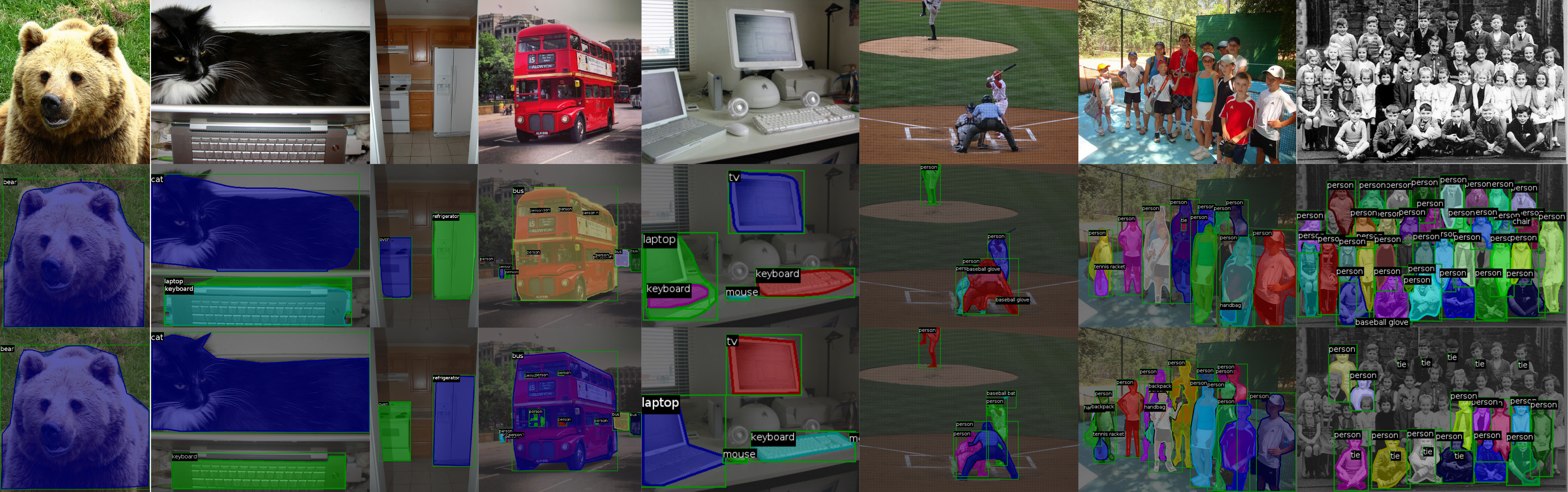}
		\caption{Some examples of semantic segmentation on MS COCO database (validation set). Each column has three images: the input image ($1^{st}$ row), our CRLS ($2^{nd}$ row) and the ground truth ($3^{rd}$ row). (Best viewed in color)}
	\label{fig:coco2015}    
\end{figure*}

The PASCAL VOC  2012 \cite{pascal-voc-2012} and the MS COCO 2014 \cite{MSCOCO} are two commonly used databases for evaluating semantic segmentation. The PASCAL VOC 2012 training and validation set has 11,530 images containing 27,450 bounding box annotated objects in 20 categories. The MS COCO training set contains about 80,000 images consisting of $\sim500,000$ annotated objects of 80 classes in total. COCO  is a more challenging dataset as it contains objects in a wide range of scales from small ($< 32^2$) to large ($> 96^2$) objects.

\begin{table}[!t]
\centering
\caption{Quantitative results and comparisons against existing CNN-based semantic segmentation methods on on the PASCAL VOC 2012 validation set.}
\label{tb:voc_test}
\begin{tabular}{lccc}
\toprule[\headrulewidth]
 \textbf{Methods} & \textbf{mAP$^r$@.5} & \textbf{mAP$^r$@.7} &  \textbf{Time (s)} \\  \midrule 

SDS (AlexNet)    	&	49.7$\%$ &	25.3$\%$ & 48\\ 
Hypercolumn  &	60.0$\%$	&	40.4$\%$    & $>$80    \\ 
CFM &	60.7$\%$	&	39.6$\%$    & 32    \\ 
MNC &	63.5$\%$	&	41.5$\%$    & \textbf{0.36}  \\ 
CRLS (Ours) &	\textbf{66.7}$\%$	&	\textbf{44.6}$\%$    & 0.54   \\
\bottomrule[\headrulewidth]
\end{tabular}
\end{table}

\begin{table}[!t]
\centering
\caption{Quantitative results and comparisons against existing CNN-based semantic segmentation methods on on the MS COCO 2014 database}
\label{tb:coco_test}
\begin{tabular}{lcc}
\toprule[\headrulewidth]
\textbf{Methods} & \textbf{mAP$^r$@[.5:.95]} & \textbf{mAP$^r$@.5} \\ \midrule 
MNC &	19.5$\%$	&	39.7$\%$ \\ 
CRLS (Ours) & \textbf{20.5}$\%$ & \textbf{40.1}$\%$ \\ 
\bottomrule[\headrulewidth]
\end{tabular}
\end{table}

We demonstrate our proposed approach on PASCAL VOC  2012. We follow the same protocols used in recent papers \cite{ Hariharan2014_SDS, Hariharan2015_Hypercolumn, Dai2015_CFM, Dai_MNC_2016} for evaluating semantic segmentation. The models are trained on the PASCAL VOC 2012 training set, and evaluated on the validation set.
The end-to-end CRLS network is trained using the ImageNet pre-trained VGG-16 model.
Results are reported on the metrics commonly used in recent semantic object segmentation papers \cite{Dai2015_CFM, Hariharan2014_SDS, Hariharan2015_Hypercolumn, Dai_MNC_2016}. We compute the mean average  precision ($\text{mAP}^r$) \cite{Dai_MNC_2016} to show the segmentation accuracy. It is measured by Intersection over Union (IoU) which indicates the intersection-over-union between the predicted and ground-truth pixels, averaged over all the classes.  In  PASCAL VOC,  we  evaluate $\text{mAP}^r$ with IoU at 0.5 and 0.7. In table \ref{tb:voc_test}, we compare our proposed CRLS with existing CNN-based semantic segmentation methods including SDS \cite{Hariharan2014_SDS}, Hypercolumn \cite{Hariharan2015_Hypercolumn}, CFM \cite{Dai2015_CFM} and MNC \cite{Dai_MNC_2016}. All the results of those methods are quoted from paper \cite{Dai_MNC_2016}.
Using the same testing protocol, our CRLS achieves higher $\text{mAP}^r$ at both 0.5 and 0.7 than previous methods (about $3\%$). In addition to high segmentation accuracy, the experimental results also show that our proposed CRLS gives very efficient testing time (0.54 second per image). Some examples of multi-instance object segmentation by our proposed CRLS on PASCAL VOC 2012 database are shown in Fig.~\ref{fig:voc2012}.

\textbf{CRLS on MS COCO Database}
\label{subsubsec:exp_coco}

Our proposed approach is trained on the MS COCO 2014 80k training images and evaluated on 20k images in the test set (\textit{test-dev}). We measure the performance of our method on two standard metrics which are the mean average precision ($\text{mAP}^r$) using IoU between 0.5 \& 0.95 and $\text{mAP}^r$ using IoU at 0.5 (as PASCAL VOC metrics) as shown in Table  \ref{tb:coco_test}.  Our  CRLS achieves better results than the previous method (MNC) on the COCO dataset (noted that we only compare with their VGG-16 network results).

\section{Discussion \& Conclusion}

This work has built the bridge between the pure image processing based variational LS methods and the learnable deep learning approaches. A novel contour evolution RLS approach has been proposed by employing GRU under the energy minimization of a variational LS functional. The learnable RLS model is designed as a building block that enable RLS easily incorporate into other deep learning components. Using shared convolutional features, RLS is extended to contextual RLS (CRLS) which combines detection, segmentation and classification tasks within a unified deep learning framework to address multi-instance object segmentation in the wild. 

As we have seen that successfully reformulating the optimization process of classic LS as a recurrent neural network is highly beneficial. In addition to solving the LS-based segmentation problem, it is worthwhile to attempt doing
the same for other iterative problems. It means our work not only boosts the classic LS approaches to new level of deep learning but also provide a novel view of point for iterative problems that not yet explored. 

The experimental results show that the proposed RLS method outperforms the baseline the Chan-Vese and the FCN methods on object segmentation task. Furthermore, the experiments on PASCAL VOC and MSCOCO concludes that the proposed CRLS gives competitive performance on semantic segmentation in the real world.
To the best of our knowledge, our proposed fully end-to-end RLS method and CRLS system are the first Level Set based methods able to deal with real-world challenging databases, i.e. PASCAL VOC and MS COCO, with highly competitive results. Our work potentially provides a vehicle for further studies in level set and iterative problems in future.

	\bibliography{refs}
	\bibliographystyle{icml2017}
	
\end{document}